\documentclass[letterpaper]{article}

\usepackage{natbib,alifeconf}
\usepackage{amsmath}
\usepackage{amssymb}
\usepackage{url,hyperref,cleveref}
\usepackage{booktabs}
\usepackage{multirow}

\title{Memory Is Communication:\\
The Frontier Between Remembering and Signaling}

\author{
    Yashar Talebirad$^{1}$,
    Eden Redman$^{2}$,
    Ali Parsaee$^{1}$, \and
    Osmar R. Za\"{i}ane$^{1}$ \\
    \mbox{}\\
    $^1$Alberta Machine Intelligence Institute, University of Alberta, Edmonton, Canada \\
    $^2$Network for Applied Technology, Edmonton, Canada \\
    talebira@ualberta.ca, eden@nat.ltd
}

\begin{document}
\raggedbottom
\maketitle

\begin{abstract}
A bounded agent may obtain information for a decision from its own past, from peers, or from both sources.
Retaining task-relevant history can reduce later communication, while a peer message can supply what memory lacks.
Under limits on both resources, how should an agent allocate its information budget?
Given a fixed task and decision rule, the memory and message rate pairs attaining a performance threshold form an achievable region under specified rules for using history and peer observations.
We call its efficient boundary the remembering--signaling frontier.
Across conditions where history permits the same maximum reduction in task loss, we hypothesize that a bounded agent will need less peer communication when it obtains a larger loss reduction from history.
In preliminary referential games, target repetition coincided with shorter successful messages, while predictability from a hidden cyclic rule did not shorten them.
Experiments varying memory and message rates can estimate the frontier and test this prediction across cooperative tasks.
\end{abstract}

\section{Introduction}

Cooperative agents often act from partial observations under limits on memory and communication bandwidth.
Existing protocols optimize which agents communicate, what they send, and to whom \citep{wang2020learning}, while local history can contain information a peer would otherwise transmit.
Memory has been described as communication between temporally separated selves \citep{levin2024self}, and we use \emph{memory is communication} for cases where retained history can replace peer input at fixed task performance.
When accumulated history exceeds what an agent can process for a single decision, hierarchical memory groups and compresses past information for retrieval under a finite input budget \citep{talebirad2026hierarchical}.
Continual learning faces the related problem of preserving usable experience as new data arrive \citep{parisi2019continual}.
Rate-distortion theory relates stored or transmitted bits to permitted error \citep{shannon1959coding}, and Wyner--Ziv coding shows how information already at a receiver can reduce the bits needed to meet an error target \citep{wyner1976rate}.
We formulate this allocation for a fixed task and decision rule, specifying which historical records and peer observations may be used.

\section{Framework and Hypotheses}

Fix a task $T$, including its instance distribution and loss, and let the decoder $A$ be the decision rule mapping the decision maker's current observation and supplied representations to the task output.
The source rules $\mathsf{src}$ identify the decision maker and its peers, and state which records from the decision maker's past and which peer observations may be encoded.
They also specify when these sources are available and exclude future records or other information unavailable to the decision maker.
Before evaluation, specify a family $\Pi_A(\mathsf{src})$ of allowed encoding schemes.
Each $\pi\in\Pi_A(\mathsf{src})$ describes how allowed history is retained and encoded as $M_\pi$, and how allowed peer observations are encoded as a message $C_\pi$.
Storing $M_\pi$ remotely still counts toward the memory rate, and the earlier transmission of a cached $C_\pi$ still counts toward communication.
Either encoded output may be empty, allowing a scheme to use one source, both sources, or neither.
Let $D_A(\pi;T)$ be the expected task loss when $A$ receives the decision maker's current observation, $M_\pi$, and $C_\pi$.
Let $R_m(\pi)$ and $R_c(\pi)$ be the expected encoded lengths of $M_\pi$ and $C_\pi$, measured in bits per decision.
The achievable memory--communication rate region is
\begin{equation}
\begin{aligned}
\mathcal{R}_A^{\mathrm{MC},\varepsilon}
  &= \{(b_m,b_c)\in\mathbb R_+^2:\\
  &\qquad \exists\pi\in\Pi_A(\mathsf{src}),\quad
      D_A(\pi;T)\leq\varepsilon,\\
  &\qquad R_m(\pi)\leq b_m,\quad R_c(\pi)\leq b_c\}.
\end{aligned}
\label{eq:region}
\end{equation}
The region depends on the decoder, task, source rules, and scheme family, and contains each budget pair under which an allowed scheme reaches expected loss at most $\varepsilon$.
The Pareto-efficient lower boundary contains allocations where reducing either rate would require increasing the other or exceeding the loss threshold.
We call this boundary the \emph{remembering--signaling frontier}.
Let $c_m,c_c>0$ be the respective per-bit costs of memory and communication.
Any region point minimizing $c_m b_m+c_c b_c$ lies on the frontier.

Our formulation counts the bits used by the historical representation as well as the peer message, and asks whether the loss reduction a bounded decoder obtains from history predicts the decrease in message rate at the same task loss.
On held-out task instances, the decoder is evaluated under three input conditions at rates fixed in advance.
The first includes only the decision maker's current observation, the second adds the history representation, and the third adds the peer message.
The loss reduction after adding history is the \emph{usable-history gain}, and the additional reduction after adding peer input is the \emph{usable-peer gain}.
The addition order affects both gains, so we keep the source order, rates, and loss measure fixed across comparisons.
A Bayes-optimal decoder has the lowest possible expected loss from the supplied inputs under the task distribution.
Among conditions where it gains equally from history, we hypothesize that the learned decoder will need fewer message bits to reach the target loss when its usable-history gain is larger.

The second hypothesis asks whether measurements that vary one rate at a time can predict the least-cost allocation when memory and message rates vary together.
Before mapping the full region, loss is measured as memory rate increases with no peer message, and then as message rate increases while memory rate is held at a value fixed in advance.
We predict that these two curves identify the memory--message split minimizing $c_m b_m+c_c b_c$ when both rates vary.
As targets repeat more often or states remain unchanged for longer, the least-cost split may shift from using more message bits to using more memory bits.
We call the point at which this shift occurs the \emph{crossover} and predict it from the two curves before varying both rates together.

\section{Preliminary Results}
\label{sec:preliminary}

To probe the relation between target predictability and message length, we varied the target process in a Lewis signaling game \citep{lewis1969convention} adapted from \citet{talebirad2026signals}, using sender--receiver pairs powered by \href{https://huggingface.co/google/gemma-4-31B-it}{instruction-tuned Gemma 4 31B}.
On each round, a sender saw the designated target among four rendered shapes drawn from eight color--shape--size combinations and sent $L\in\{1,2,3\}$ symbols over $\{A,B,C\}$ to a receiver, allowing $3^L$ messages, equivalent to $L\log_2 3$ bits per round.
The receiver saw the same shapes and guessed the target.
The symbols had no assigned meanings, and feedback revealed the target and outcome after each guess.
Each agent retained a private notebook and the previous twenty interactions.
For each condition and seed, separate pairs played forty rounds for $L=1,2,3$.
We recorded the smallest $L$ whose pair achieved accuracy $\geq0.85$ over the final ten rounds as $L_{\min}$ (chance $=0.25$).

In the repeat condition, the next target copied the previous target with probability $p$ and was otherwise sampled uniformly.
Across two batches of three seeds, mean $L_{\min}$ fell monotonically from $2.67$ at $p=0$ to $1.00$ at $p=0.95$.
Because episode history was never removed while learned symbol mappings were preserved, these runs do not isolate history's contribution.

\begin{table}[h!]
\centering
\scriptsize
\caption{Mean $L_{\min}$ over three seeds. Repeat uses $p=(0,0.5,0.8,0.95)$ and rotation uses $p=(0,0.5,0.85,1)$.}
\label{tab:lmin}
\begin{tabular}{lcccc}
\toprule
Target process & \multicolumn{4}{c}{$L_{\min}$ as $p$ increases} \\
\midrule
Repeat, batch 1 & 2.67 & 2.00 & 1.67 & 1.00 \\
Repeat, batch 2 & 2.67 & 2.33 & 1.67 & 1.00 \\
Shuffled rotation & 2.33 & 2.00 & 2.67 & 3.00 \\
\bottomrule
\end{tabular}
\end{table}

Under shuffled rotation, the eight targets formed a fixed, randomly ordered cycle hidden from the agents.
On each round, the sequence advanced one step with probability $p$ and otherwise jumped to a random target.
When $p=1$, the sequence completed five cycles in forty rounds, yet mean $L_{\min}$ increased from $2.33$ at $p=0$ to $3.00$ at $p=1$.

\section{Discussion}

The contrasting trends motivate our hypothesis, which remains to be tested with usable-history measurements and message-rate predictions registered in advance.
Interpretation is further limited because the target processes differed in rule complexity and in the number of distinct targets encountered, while the evidence comes from one model family, three seeds per batch, and forty interaction rounds.
These limitations motivate three tests of increasing generality.

The first uses a small state-estimation task in which the hidden state remains the same from one round to the next with a fixed probability and otherwise changes.
An agent estimates its value from a compressed record of its own noisy observations and a rate-limited message about a peer's noisy observation.
Because states and codes are finite, the Bayes-optimal frontier can be calculated exactly for the allowed schemes and compared with the learned decoder's frontier.

The second maps the tested part of the region by varying memory rate, message rate, and target predictability in the referential game.
Within each condition, all rate settings use the same target stream, and comparisons across processes match target frequencies and the Bayes-optimal gain from history.
A memory ablation removes episode history while preserving the learned symbol mappings.
Longer runs and another model family can distinguish insufficient learning time from an inability to use the rotation reliably.

The third test repeats these rate experiments on distributed tasks with local views.
AgentsNet provides coloring, maximal matching, vertex cover, leader election, and consensus \citep{grotschla2025agentsnet}.
LoopBench provides repeated graph coloring in which observable actions may convey information \citep{talebirad2025loopbench}.
We preregister a subset containing local and global coordination tasks, then vary how long private inputs or local states remain unchanged while holding the communication graph and task objective fixed.
To compare tasks, we charge memory, explicit messages, and information conveyed by observable actions against the same bit budget.
For each task and model family, we use measurements that vary one rate at a time to predict the crossover before varying both rates together.

\section*{Acknowledgments}
This research was supported by the Alberta Machine Intelligence Institute (Amii) and the Canada CIFAR AI Chairs Program.
We also thank the Network for Applied Technology (NAT) for its support.
The authors used Claude (Anthropic) and Codex (OpenAI) to assist with code development and manuscript editing.
All AI-assisted outputs were reviewed and verified by the authors, who take full responsibility for the work.

\footnotesize
\bibliographystyle{apalike}
\bibliography{references}

@Book{lewis1969convention,
  author    = {Lewis, D.},
  title     = {Convention: A Philosophical Study},
  publisher = {Harvard University Press},
  year      = {1969}
}

@InProceedings{shannon1959coding,
  author    = {Shannon, C. E.},
  title     = {Coding Theorems for a Discrete Source with a Fidelity Criterion},
  booktitle = {IRE National Convention Record},
  year      = {1959},
  volume    = {4},
  pages     = {142--163}
}

@Article{wyner1976rate,
  author  = {Wyner, A. and Ziv, J.},
  title   = {The Rate-Distortion Function for Source Coding with Side
             Information at the Decoder},
  journal = {IEEE Transactions on Information Theory},
  year    = {1976},
  volume  = {22},
  number  = {1},
  pages   = {1--10},
  doi     = {10.1109/TIT.1976.1055508}
}

@InProceedings{wang2020learning,
  author    = {Wang, Rundong and He, Xu and Yu, Runsheng and Qiu, Wei and
               An, Bo and Rabinovich, Zinovi},
  title     = {Learning Efficient Multi-Agent Communication: An Information
               Bottleneck Approach},
  booktitle = {Proceedings of the 37th International Conference on Machine
               Learning},
  series    = {Proceedings of Machine Learning Research},
  volume    = {119},
  pages     = {9908--9918},
  year      = {2020},
  url       = {https://proceedings.mlr.press/v119/wang20i.html}
}

@Article{levin2024self,
  author  = {Levin, Michael},
  title   = {Self-Improvising Memory: A Perspective on Memories as
             Agential, Dynamically Reinterpreting Cognitive Glue},
  journal = {Entropy},
  year    = {2024},
  volume  = {26},
  number  = {6},
  pages   = {481},
  doi     = {10.3390/e26060481}
}

@Article{grotschla2025agentsnet,
  author  = {Gr{\"o}tschla, Florian and M{\"u}ller, Luis and
             T{\"o}nshoff, Jan and Galkin, Mikhail and Perozzi, Bryan},
  title   = {{AgentsNet}: Coordination and Collaborative Reasoning in
             Multi-Agent {LLM}s},
  journal = {arXiv preprint arXiv:2507.08616},
  year    = {2025},
  url     = {https://arxiv.org/abs/2507.08616}
}

@Article{talebirad2025loopbench,
  author  = {Parsaee, Ali and Talebirad, Yashar and Szepesv{\'a}ri, Csongor and
             Ohal, Vishwajeet and Redman, Eden},
  title   = {{LoopBench}: Discovering Emergent Symmetry Breaking
             Strategies with {LLM} Swarms},
  journal = {arXiv preprint arXiv:2512.13713},
  year    = {2025},
  url     = {https://arxiv.org/abs/2512.13713}
}

@Article{talebirad2026signals,
  author  = {Talebirad, Yashar and Redman, Eden and Parsaee, Ali and
             Za\"{i}ane, Osmar R.},
  title   = {From Signals to Structure: How Memory Architecture
             Drives Language Emergence in {LLM} Agents},
  journal = {arXiv preprint arXiv:2607.00233},
  year    = {2026},
  doi     = {10.48550/arXiv.2607.00233},
  note    = {Accepted at the 2026 Conference on Artificial Life}
}

@article{parisi2019continual,
  title = {Continual Lifelong Learning with Neural Networks: A Review},
  author = {Parisi, German I. and Kemker, Ronald and Part, Jose L. and Kanan, Christopher and Wermter, Stefan},
  journal = {Neural Networks},
  volume = {113},
  pages = {54--71},
  year = {2019},
  doi = {10.1016/j.neunet.2019.01.012},
  url = {https://doi.org/10.1016/j.neunet.2019.01.012}
}

@misc{talebirad2026hierarchical,
  title = {Toward a Theory of Hierarchical Memory for Language Agents},
  author = {Talebirad, Yashar and Parsaee, Ali and Szepesv\'{a}ri, Csongor Y. and Nadiri, Amirhossein and Za\"{i}ane, Osmar R.},
  year = {2026},
  note = {ICLR 2026 Workshop on Memory for LLM-Based Agentic Systems},
  url = {https://arxiv.org/abs/2603.21564}
}

\end{document}